\documentclass[journal]{vgtc}                % final (journal style)
\ifpdf%                                % if we use pdflatex
  \pdfoutput=1\relax                   % create PDFs from pdfLaTeX
  \usepackage{graphicx}                % allow us to embed graphics files
  \DeclareGraphicsExtensions{.pdf,.png,.jpg,.jpeg} % for pdflatex we expect .pdf, .png, or .jpg files
\else%                                 % else we use pure latex
  \usepackage{graphicx}                % allow us to embed graphics files
  \DeclareGraphicsExtensions{.eps}     % for pure latex we expect eps files
\fi%

\graphicspath{{figures/}{pictures/}{images/}{./}} % where to search for the images

\usepackage{microtype}                 % use micro-typography (slightly more compact, better to read)
\PassOptionsToPackage{warn}{textcomp}  % to address font issues with \textrightarrow
\usepackage{textcomp}                  % use better special symbols
\usepackage{mathptmx}                  % use matching math font
\usepackage{times}                     % we use Times as the main font
\usepackage{cite}                      % needed to automatically sort the references
\usepackage{tabu}                      % only used for the table example
\usepackage{booktabs}
\usepackage[usenames, dvipsnames]{color}
\usepackage{csquotes}

\usepackage{tabularx}
\usepackage{enumitem}

\newcommand{\say}[1]{\textquote{#1}}

\newcommand{\tool}{\textsc{Seq2Seq-Vis} }

\newcommand{\seqseq}{seq2seq}

\onlineid{0}

\vgtccategory{Research}
\vgtcpapertype{please specify}

\title{\tool: A Visual Debugging Tool for\\ Sequence-to-Sequence Models}

\author{Hendrik Strobelt*, Sebastian Gehrmann*, Michael Behrisch, Adam Perer,
Hanspeter Pfister, Alexander M. Rush}
\authorfooter{
\item 
(*) indicates equal contribution\\
\item
 H. Strobelt is with IBM Research and MIT-IBM Watson AI Lab.
\item
 S. Gehrmann and A. Rush are with the Harvard NLP group.
\item
 M. Behrisch and H. Pfister are with the Harvard Visual Computing group.
\item 
 A. Perer is with Carnegie Mellon University.
}

\shortauthortitle{Biv \MakeLowercase{\textit{et al.}}: Global Illumination for Fun and Profit}
\abstract{Neural sequence-to-sequence models have proven to be
accurate and robust for many sequence prediction tasks, and have
become the standard approach for automatic translation of text. The
models work with a five-stage blackbox pipeline that begins with encoding a
source sequence to a vector space and then decoding out to a new
target sequence. This process is now standard, but like many deep
learning methods remains quite difficult to understand or debug.  In
this work, we present a visual analysis tool
that allows interaction and "what if"-style exploration of trained sequence-to-sequence models through each stage of the translation process. The aim is to identify
which patterns have been learned, to detect model errors, and to probe the model with
counterfactual scenario. We demonstrate the utility of our tool through several real-world  sequence-to-sequence use cases on large-scale models.

} % end of abstract
\CCScatlist{ % not used in journal version
 \CCScat{K.6.1}{Management of Computing and Information Systems}%
{Project and People Management}{Life Cycle};
 \CCScat{K.7.m}{The Computing Profession}{Miscellaneous}{Ethics}
}

\teaser{
  \centering
  \includegraphics[]{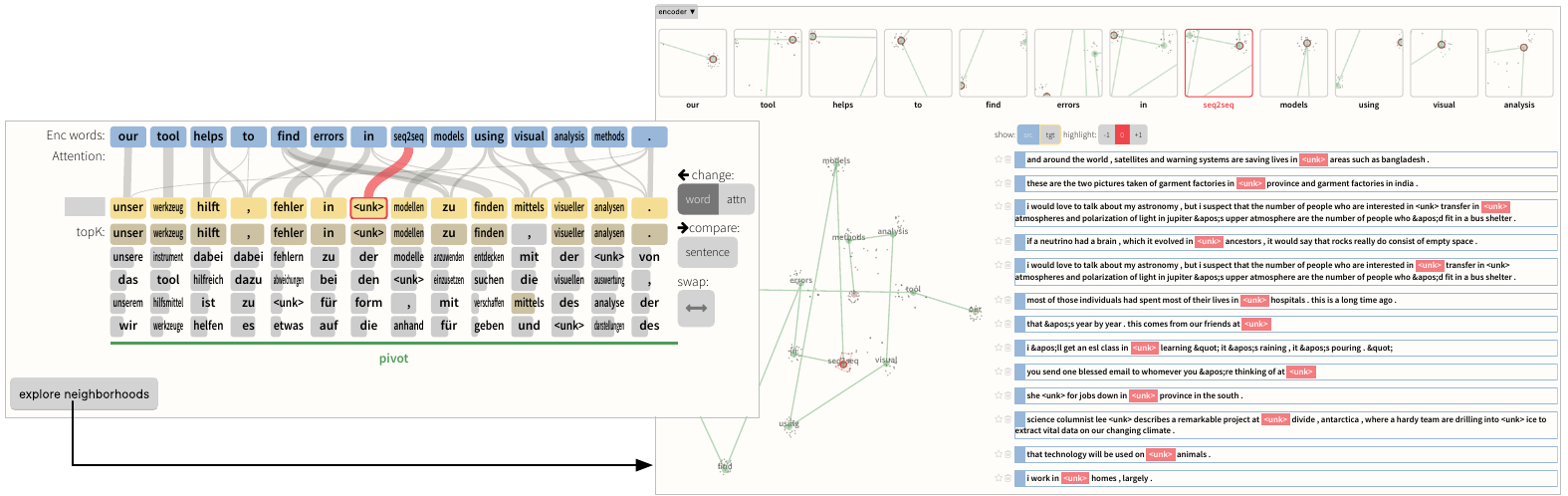}
  \caption{Example of \texttt{Seq2Seq-Vis}. In the translation view (left), the source sequence \textit{``our tool helps to find errors in seq2seq models using visual analysis methods.''} is translated into a German sentence. The word \textit{``seq2seq''} has correct attention between encoder and decoder (red highlight) but is not part of the language dictionary. When investigating the encoder neighborhoods (right), the user sees that \textit{``seq2seq''} is close to other unknown words \textit{``$\langle$unk$\rangle$''}. The buttons enable user interactions for deeper analysis.}
	\label{fig:teaser_small}
}

\vgtcinsertpkg

\begin{document}

%% The ``\maketitle'' command must be the first command after the
%% ``\begin{document}'' command. It prepares and prints the title block.

%% the only exception to this rule is the \firstsection command
\firstsection{Introduction}

\maketitle

% Deep learning has become more and more a standard tool for
% solving machine learning tasks across a wide variety of data
% domains. ... ARGH
% \ap{These first sentences are very NLP-centric.  Should we make it more broader?  Also, can we mention Deep Learning up front so we make it clear that we are working on what VAST thinks is 'cool stuff'.}
% \hs{see below -- I merged with intro to Sec 2 -- idea from MB }

Deep learning approaches based on neural networks have shown
significant performance improvements on many artificial intelligence
tasks.  However, the complex structure of these networks often makes
it difficult to provide explanations for their predictions.
Attention-based \emph{sequence-to-sequence models} (seq2seq)~\cite{sutskever2014sequence,bahdanau2014neural}, also known as
encoder-decoder models, are representative of this trend. Seq2seq
models have shown state-of-the-art performance in a broad range of
applications such as machine translation, natural language generation,
image captioning, and summarization. Recent results show that these
models exhibit human-level performance in machine translation for certain
important domains~\cite{wu2016google,awadalla2018achieving}.

% Recently, variants of \seqseq~models featuring
% convolutional and attention-only components\cite{}, however these
% extensions even further complicate the \seqseq~ architecture.

%Recent advances in natural language processing and
%generation have been driven by the development and refinement of
%\textit{sequence-to-sequence} models. These models have supported
%major improvements in the benchmark tasks of machine translation
%leading to human-level performance on some important domains \cite{}
%as well as advances in related areas like summarization, dialogue
%response generation, image captioning, and data-to-text generation.

Seq2seq~models are powerful because they provide an effective
supervised approach for processing and predicting sequences without
requiring manual specification of the relationships between source and
target sequences. Using a single model, these systems learn to do
reordering, transformation, compression, or expansion of a source
sequence to an output target sequence. These modifications are
performed using a large internal state representation first encodes and
then decodes the source sequence. With enough data, these models provide a general purpose mechanism for learning to predict sequences.

While the impact of seq2seq models has been clear, the added
complexity and uncertainty of deep learning based models raises
issues.  These models act as black-boxes during prediction,
making it difficult to track the source of mistakes.  The high-dimensional internal representations make it difficult to analyze the model as it transforms the data.  While this property is shared across deep learning, mistakes involving language are often very apparent to human readers.  For instance, a widely publicized incident resulted from a
seq2seq translation system mistakenly translating ``good morning''
into ``attack them'' leading to a wrongful arrest~\cite{hern2017}.
Common but worrying failures in \seqseq~models include
machine translation systems greatly mistranslating a sentence, image
captioning systems yielding an incorrect caption, or speech
recognition systems producing an incorrect transcript.

Ideally, model developers would understand and trust the results of their systems,
but currently, this goal is out of reach. In the
meantime, the visual analytics community can contribute to
this crucial challenge of better surfacing the
mistakes of seq2seq~systems in a general and reproducible way. We propose \tool, a visual analytics tool that satisfies this
criteria by providing support for the following three goals:
\begin{itemize} %[leftmargin=*,noitemsep,nolistsep]
\item \textbf{Examine Model Decisions:} \tool allows users to
  understand, describe, and externalize model errors for each
  stage of the seq2seq~pipeline.
\item \textbf{Connect Decisions to Samples:} \tool describes the origin of
  a seq2seq~model's decisions by relating internal states to relevant training samples.
\item \textbf{Test Alternative Decisions:} \tool facilitates model interventions by 
making it easy to manipulate of model internals and conduct "what if" explorations.
\end{itemize}

The full system is shown in Figure~\ref{fig:teaser_small} (or larger in \autoref{fig:overview}). It integrates
visualizations for the components of the model (Fig~\ref{fig:teaser_small}
left) with internal representations from specific examples
(Fig~\ref{fig:teaser_small} middle) and nearest-neighbor lookups over a
large offline corpus of precomputed examples (Fig~\ref{fig:teaser_small}
right).

We begin in \autoref{sec:s2smodelsattention} by introducing
important background and notation to formalize our
overall goal of seq2seq~model debuggers. In \autoref{uc_dark} we present a guiding example illustrating how a typical model understanding- and debugging session looks like for an analyst. The subsequent \say{Goals and  Task} Section \ref{sec:goals-and-tasks}  enumerates the goals and procedures for building a seq2seq
debugger. Based on these guidelines, \autoref{sec:design} and
\autoref{sec:implementation} introduce our visual and
implementation design choices. \autoref{sec:use_cases} highlights in
three further real-world use cases for how \tool guides the user through the visual analysis process. \autoref{sec:related-work} puts
these contributions in the context of related work for this research
domain and \autoref{sec:conclusion} presents future work and
reflections.

% This
% example illustrates that there is an urgent need for systems that can
% detect systematic flaws in what neural models learn.
% However there
% are also clear downsides of using a

% Modern techniques for natural language processing and generation
% have

% Hen (1st) + HP (2nd)

% \textbf{Theme: It's complicated and new :)}

% SG: Use if you want, I like the facebook example

% Neural models are complex finite state machines
% that are trained to embed language in a high-dimensional latent
% space. Thus, a sequence of inputs lead the model to generate a
% trajectory through this space, which is comparable to a finite state
% machine moving from state to state.  Following this metaphor, a neural
% model predicts by selecting the next word that has the most similar
% latent state in the training data. In this work, we show that this is
% true in most cases. Therefore, a model fails when a particular set of
% inputs leads it to stray away from the \emph{correct} state it should
% move to, of if the model receives an unanticipated input that it has
% not seen in the training data.

\begin{figure}[ht]
\centering
\includegraphics[width=\columnwidth]{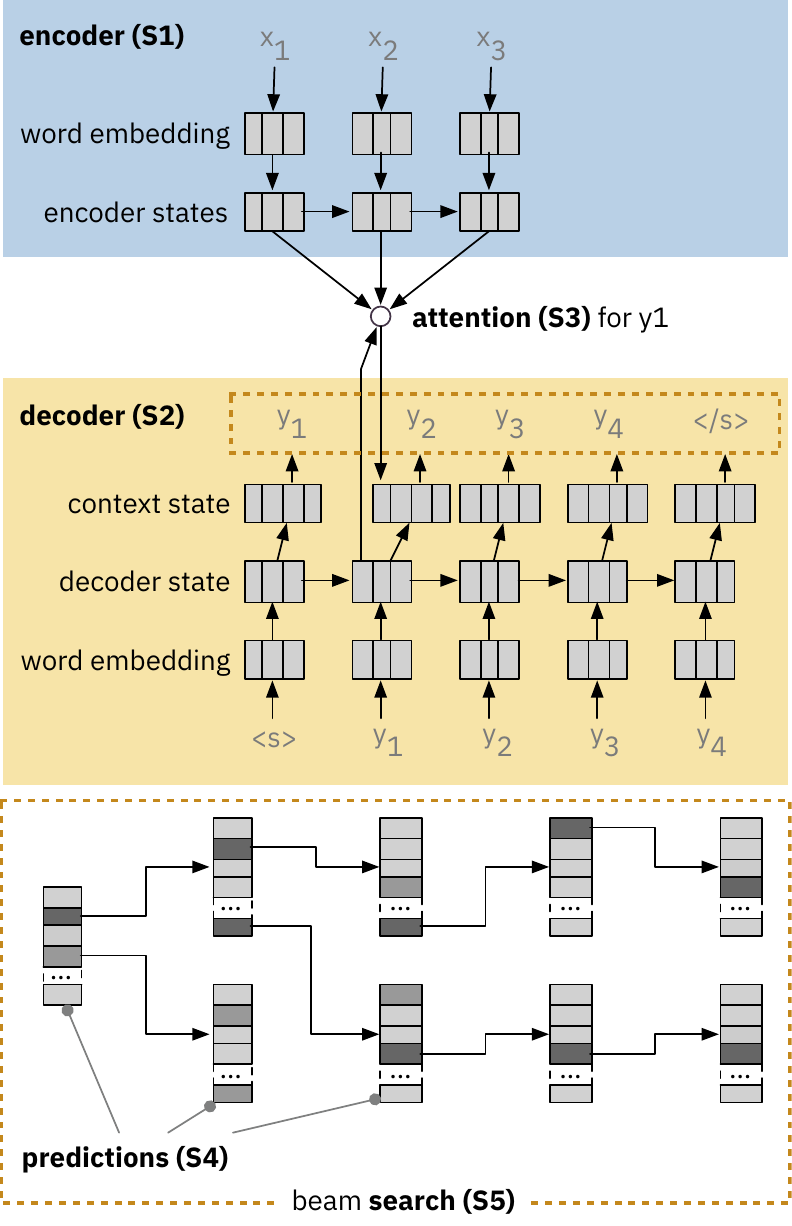}
\caption{Five stages in translating a source to target sequence: (S1) encoding the source sequence into latent vectors, (S2) decoding to generate target latent vectors, (S3) attend between encoder and decoder, (S4) predict word probabilities at each time step, and (S5) search for a best complete translation (beam search).}
\label{fig:s2smodel}
\end{figure}

\section{Sequence-to-Sequence Models and Attention}
\label{sec:s2smodelsattention}

We begin with a formal taxonomy of seq2seq models that will inform our
visual analytics approach.  Throughout this work, for brevity and
clarity, we will consider the running example of automatic translation
from one language to another. We use the sequence notation
$\mathbf{x}_{1:S}$ to represent an $S$-word sentence in a source
language, and $\mathbf{y}_{1:T}$ to represent a $T$-word sentence in
a target language.  Seq2seq models perform translation in
a left-to-right manner, one target word at a time, until a special stop token is
generated, which ends the translation. 

We break down the translation process of seq2seq models into
five stages: (S1) \textit{encode} the source sentence, (S2)
\textit{decode} the current target words, (S3) \textit{attend} to the encoded
source, (S4) \textit{predict} the next target word, and (S5)
\textit{search} for the best complete translation. Note that some
systems use a slightly different order, but most adhere roughly to
this setup.  \autoref{fig:s2smodel} provides a structural overview of
these five stages.

\noindent\textbf{Encoder (S1):} Encoding uses a deep neural network to convert a sequence of source words $\mathbf{x}_{1:S}$
into a sequence of vectors
$\mathbf{x}'_{1:S}$. Each vector in the sequence
$\mathbf{x}'_{s}$ roughly represents one word
$\mathbf{x}_s$ but also takes into account the surrounding
words, both preceding and succeeding, that may determine its contextual meaning. This encoding is
typically done using a recurrent neural network (RNN) or a long
short-term memory network (LSTM), however recently non-RNN-based
methods such as convolutional neural networks (CNN) \cite{gehring2017convs2s, gehring2016convenc} and Transformer \cite{vaswani2017attention, liu2018generating} have also been
employed. Our approach supports all types of encoding methods.

\noindent\textbf{Decoder (S2):} The decoding process is analogous to encoding, and takes the sequence of previously generated target words $\mathbf{y}_{1:t}$ and
converts them to a sequence of latent vectors
$\mathbf{y}'_{1:t}$. Each vector represents the state of
the sentence up to and including word $\mathbf{y}_t$. This provides a
similar contextual representation as in the encoder, but is only based
on previous words. Upon producing a new word, the prediction is used as input to the decoder.

\noindent\textbf{Attention (S3):} The attention component matches encoder hidden states and decoder hidden states.
For each $\mathbf{y}'_{t}$ we consider which encoder
states $\mathbf{x}'_{s}$ are relevant to the next
prediction. In some similar language pairs, like French and Spanish,
the words often align in order, e.g., the fourth French word matches the
fourth Spanish word.  However for languages such as English and
Chinese, the matching might be quite far away. Instead of using
absolute position, attention compares the word representations to find
which source position to translate. Attention forms a
distribution based on the dot product between vectors
$\mathbf{x}'_{s} \cdot \mathbf{y}'_{t}$.
We call this value $a_{s, t}$, and it indicates how closely
the source and target positions match.

\noindent\textbf{Prediction (S4):} The prediction step produces a
multi-class distribution over all the words of the target language -- words that are more likely to come next have higher probability.
This problem takes two factors into account: the current
decoder state $\mathbf{y}'_t$ and the encoder states
weighted by attention,
known as the context vector coming from S3. These two are combined
to predict a distribution over the next word
$p(\mathbf{y}_{t+1} | \mathbf{x}_{1:S}, \mathbf{y}_{1:t})$.

\noindent\textbf{Search (S5):}
To actually produce a translation, these previous steps are combined
into a search procedure. Beam search is a variant of standard tree search 
that aims to efficiently explore the space of translations. 
The deep learning component of Seq2seq models  predicts the probability of all next words, 
given a prefix. While one could simply
take the highest probability word at each time step, it is possible
that this choice will lead down a bad path (for instance, first picking the word "an"
and then wanting a word starting with a consonant).
\textit{Beam search} instead pursues several possible \textit{hypothesis} translations
each time step. It does so by building a tree comprising the top $K$-
hypothesis translations. At each point, all next words are generated for each. Of these, only the most likely $K$ are preserved. Once all $K$ beams have terminated by generating the stop token, the final prediction is the translation with the highest score.  

Each stage of the process is crucial for effective translation, and it
is hard to separate them. However, the model does preserve some
separations of concerns. The decoder (S2) and encoder (S1) primarily
work with their respective language, and manage the change in hidden
representations over time. Attention (S3) provides a link between the
two representations and connects them during training. Prediction
(S4) combines the current decoder state with the information moving
through the attention. Finally, search (S5) combines these
with a global score table.  These five stages provide the foundation
for our visual analytics system.

\section{Motivating Case Study: Debugging Translation}
\label{uc_dark}

To motivate the need for our contributions, we present a representative case study. Further case studies are discussed in
\autoref{sec:use_cases}. This case study involves a model trainer (see
\cite{strobelt2018lstmvis}) who is building a German-to-English
translation model (our model is trained on the small-sized IWSLT '14
dataset~\cite{mauro2012wit3}).

%Before describing the Design (\autoref{sec:design}) of \texttt{Seq2Seq-Vis} in
%detail, we first demonstrate the interplay of core components through
%a tangible example.

% and the
% interaction with a translation model by an introductory example.  For
% brevity, we leave algorithmic details for later sections.
%For this case study, we use a production-ready German-to-English
%translation model trained for the WMT 2017 task \cite{}.

\begin{figure}[ht]
\centering
	\includegraphics[width=\columnwidth]{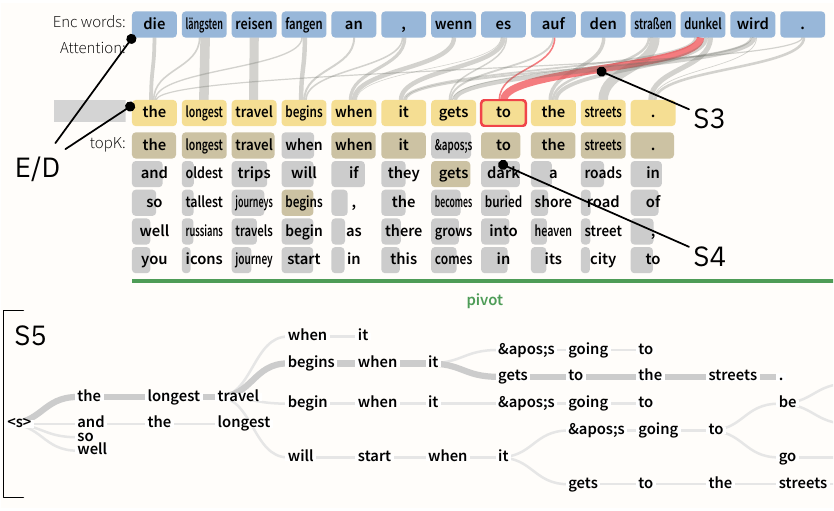}
	\caption{Hypotheses: Attention (S3), Prediction (S4), or Beam Search (S5) Error -- encoder words and decoder words (E/D), Attention (S3), top k predictions for each time step in decoder (S4), and beam search tree (S5) }
	\label{fig:uc_dark_03}
\end{figure}

The user begins by seeing that a specific example was mistranslated in
a production setting. She finds the source sentence: \textit{Die l\"angsten Reisen fangen an, wenn es auf den Stra\ss en dunkel
  wird.}\footnote{The closing quote of the book `Kant' from German
  author J\"org Fauser, who is attributed as being a forerunner of
  German underground literature.}  This sentence should have been
 translated to: \textit{The longest journeys begin, when it gets
  dark in the streets.}  She notices that the model produces the
mistranslation: \textit{The longest journey begins, when it gets to the
  streets.}  \autoref{fig:uc_dark_03}(E/D) shows the tokenized input
sentence in blue and the corresponding translation of the model in
yellow (on the top).  The user observes that the model does not translate the word \textit{dunkel} into \textit{dark}.

This mistake exemplifies several goals that motivated the development of \texttt{Seq2Seq-Vis}. The
user would like to examine the system's decisions, connect to training
examples, and test possible changes. As described in
\autoref{sec:s2smodelsattention}, these goals apply to all five model
stages: encoder, decoder, attention, prediction, and search.

%\ap{Instead of using 'we' throughout the section, consider adopting the persona of a data scientist. e.g. \emph{The data scientist wishes to X,Y,Z.}}

\begin{figure}[ht]
\centering
	\includegraphics[width=\columnwidth]{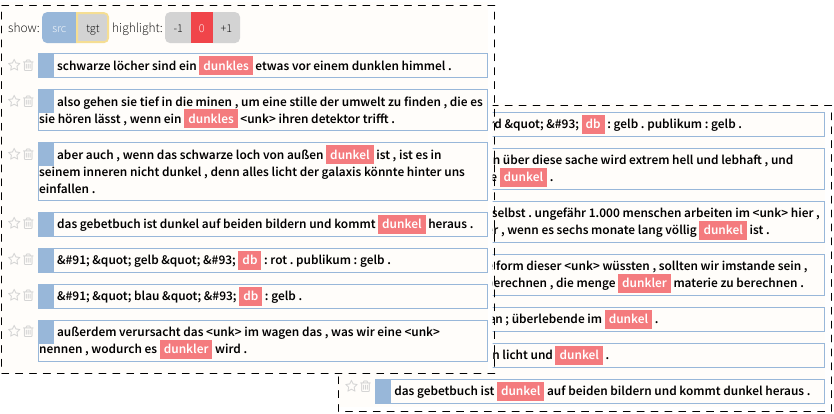}
	\caption{Hypothesis: Encoder~(S1) Error -- nearest neighbors of encoder state for \textit{dunkel}.}
	\label{fig:uc_dark_01}
\end{figure}

\noindent \textbf{Hypothesis: Encoder~(S1) Error?}
\texttt{Seq2Seq-Vis} lets the user examine similar encoder states for
any example. Throughout, we will use the term \emph{neighborhood} to refer to the twenty closest states in vector space from training data.
%To facilitate exploration, a trace is shown of projected neighbors
%(\autoref{fig:uc_dark_all}.H1 left) of all sentence encoder states.
\tool displays the nearest neighbor sentences for a specific encoder
state as red highlights in a list of training set examples.
\autoref{fig:uc_dark_01} shows that the nearest neighbors for
\textit{dunkel} match similar uses of the word.  The majority seem to
express variations of \textit{dunkel}.  The few exceptions,
e.g., \textit{db}, are artifacts
that can motivate corrections of the training
data %(for the \textit{db} case)
or trigger further investigation. Overall, the encoder seems to
perform well in this case.

%\ap{How will readers know how to interpret \autoref{fig:uc_dark_all}.H1 since it hasn't been explained yet?  You either need to a) explain here, b) forward-reference (bad!) or c) switch the order of these sections.  Could you integrate Section 4 and Section 5, so you explain the UI through the lens of this example?  That might be the cleanest approach from a reader's POV.}

\begin{figure}[ht]
\centering
	\includegraphics[width=\columnwidth]{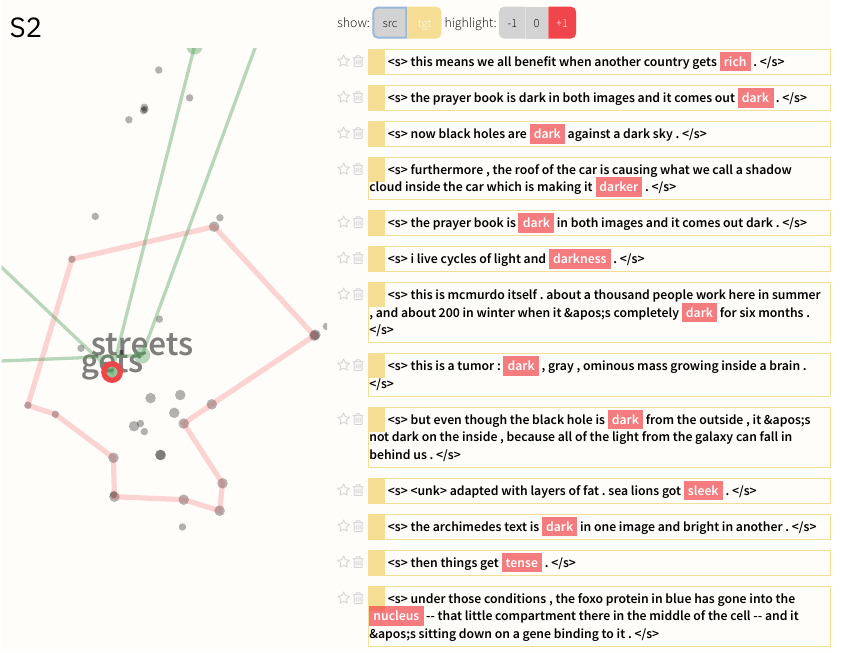}
	\caption{Hypothesis: Decoder~(S2) Error -- nearest neighbors of decoder state for \textit{gets} and \textit{streets}, which are close in projection space.}
	\label{fig:uc_dark_02}
\end{figure}

\noindent \textbf{Hypothesis: Decoder~(S2) Error?}  Similarly, the
user can apply \tool to investigate the neighborhood of
decoder states produced at times $t$ and $t+1$
(\autoref{fig:uc_dark_02}). In addition to the neighbor list, it gives
a projection view that depicts all decoder states for the current
translation and all their neighbors in a 2D plane.  The analyst
observes that the decoder states produced by \textit{gets} and
\textit{streets} are in proximity and share neighbors.  Since
these states are indicative for the next word we can switch
the highlight one text position to the right (+1) and observe that the
decoder states at \textit{gets} and \textit{streets} support
producing \textit{dark}, \textit{darker}, or \textit{darkness}. Thus,
the decoder state does not seem very likely as the cause of the error.

% Similar to the
% encoder side, we can refer the other semantic embeddings (like
% \textit{albania} or \textit{nucleus}) to further investigation or
% removal from training data. But conclusively, H2 does not seem very
% likely to be cause of error.

\noindent \textbf{Hypothesis: Attention~(S3) Error?} Since both
encoder and decoder are working, another possible issue is that the
attention may not focus on the corresponding source token
\textit{dunkel}. The previous hypothesis testing revealed that well-supported positions for adding \textit{dark} are after \textit{gets}
or \textit{streets}. This matches human intuition, as we can imagine
the following sentences being valid translations: \textit{The longest
  travels begin when it gets \textit{dark} in the streets.} or
\textit{The longest travels begin when it gets to the streets turning
  \textit{dark}.}  In \autoref{fig:uc_dark_03}(S3) our analyst can
observe that the highlighted connection following \textit{get} to the
correct next word \textit{dunkel} is very strong. The
connection width indicates that the attention weight is very high with
the correct word. Therefore, the user can assume that the attention is well set for predicting \textit{dark} in this position.  The hypothesis for
error in S3 can be rejected with high probability.

\noindent \textbf{Hypothesis: Prediction~(S4) Error?} The combination
of decoder state and attention is used to compute the probability of
the next word. It may be that an error occurs in this decision,
leading to a poor probability of the word \textit{dark}. The tool
shows the most likely next words and their probabilities in
\autoref{fig:uc_dark_03}(S4). Here, our analyst can see that the model mistakenly assigns a higher probability to \textit{to} than
\textit{dark}. However, both options are very close in probability, indicating that the model is quite uncertain and almost equally split
between the two choices. These local mistakes should be automatically
fixed by the beam search, because the correct choice \textit{dark} leads to a globally more likely sentence.

\noindent \textbf{Hypothesis: Search~(S5) Error?}  Having eliminated
all other possible issues, the problem is likely to be a search
error. The user can investigate the entire beam search tree in
\autoref{fig:uc_dark_03}(S5), which shows the top $K$ considered options at each prediction step. In this case, the analyst finds that \textit{dark} is never considered within the search. Since the previous test showed that \textit{to} is only minimally more likely than \textit{dark}, a larger $K$ would probably have lead to the model considering \textit{dark} as the next best option. 
We therefore conclude that this local bottleneck of a too narrow beam search is the most likely error case. The analyst has identified a search error, where the approximations made by beam search cut off the better global option in favor of a worse local choice.

% \textbf{Hypothesis H4 - The beam search is not working optimally.}
% We observe the beam search tree in \autoref{fig:uc_dark_all}.H4. For the current prediction we see the beam of highest support highlighted. But in none of the alternative beams we see the semantics of dark being represented. We can conclude that this might be the most likely hypothesis.

%\begin{figure}
%\centering
%	\includegraphics[width=\columnwidth]{figures/uc_dark_04}
%	\caption{...}
%	\label{fig:uc_dark_04}
%\end{figure}

\begin{figure}[ht]
\centering
	\includegraphics[width=\columnwidth]{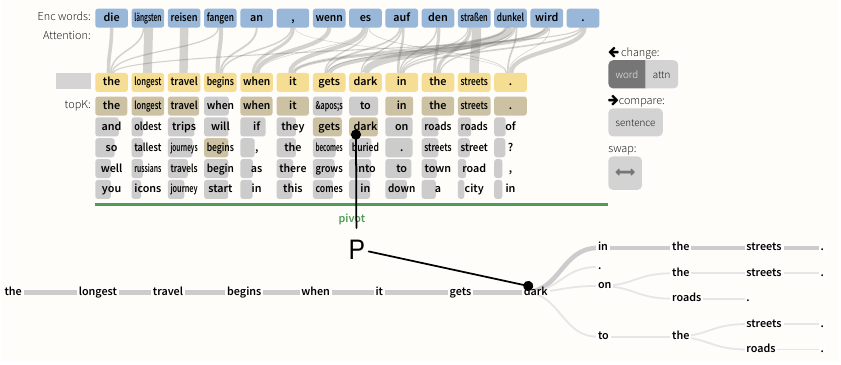}
	\caption{Testing a fix -- by clicking on the correct word \textit{dark} in the predicted top-K, the beam search is forced on a specific path (P) which leads to the correct prediction.}
	\label{fig:uc_dark_04}
\end{figure}

 \noindent \textbf{Exploring Solutions.} 
 When observing the $K$-best predictions for the position of \textit{to}, the analyst sees that \textit{dark} and \textit{to} are close in
 probability (\autoref{fig:uc_dark_03}(S4)). To investigate whether the model would produce the correct answer if it had considered \textit{dark}, \tool allows the user to evaluate a case-specific fix. The analyst can test this counterfactual, what would have happened if she had forced the translation to use
 \textit{dark} at this critical position? By clicking on \textit{dark}
 she can produce this probe (shown in \autoref{fig:uc_dark_04}), which yields the correct translation. The user can now describe the most
 likely cause of error (search error) and a local fix to the problem
 (forced search to include dark). The analyst can now add this case to
 a list of well-described bugs for the model and later consider
 a global fix.

\section{Goals and Tasks}\label{sec:goals-and-tasks}

% \begin{table}
% \centering
% \caption{steps of debugging and role of \texttt{Seq2Seq-Vis}. \hs{keep or trash?}}
% \label{my-label}
% %	\toprule
% %\rowcolors{2}{gray!25}{white}
% 	\begin{tabularx}{.9\columnwidth}{l|X}
% 		\thead{step} & \thead{approach} \\
% 	\midrule
% 	  Find critical samples &  rank test set by performance score or user feedback  \\
% 	  Where did the model fail? & \texttt{Seq2Seq-Vis} (G1)  \\
% 	  Which rule caused failure? & \texttt{Seq2Seq-Vis} (G2)  \\
% 	  Test possible fixes & \texttt{Seq2Seq-Vis} (G3)  \\
% 	  Fix bug permanently & future work in algorithms\\
% %	\bottomrule
% 	\end{tabularx}
% \end{table}

% We use the running example as specific instance to motivate the following
% generalization of domain goals and inferred visualization and interaction tasks.

We now step back from this specific instance and consider a common
deployment cycle for a deep learning model such as seq2seq.  First, a
model is trained on a task with a possibly new data set, and then
evaluated with a standard metric. The model performs well in aggregate
and the stakeholders decide to deploy it. However, for a certain
subset of examples there exist non-trivial failures. These may be noticed by users, or, in the case of translation, by post-editors who correct the output of the system.
While the model itself is still useful, these examples might be
significantly problematic as to cause alarm.

% When these errors occur,
% there is very little that can be done currently, besides removing the
% model temporarily, or substituting a simpler model in its place, that
% may be more interpretable but often with lower accuracy.

Although these failures can occur in any system, this issue was much less
problematic in previous generations of AI systems. For instance when
using rule-based techniques, a user can explore the provenance of a
decision through rules activated for a given output. If there is a
mistake in the system, an analyst can 1) identify which rule misfired,
2) see which previous examples motivated the inclusion of the rule, and
3) experiment with alternative instances to confirm this behavior.

% This
% feedback loop provides a route for system diagnostics that we want to
% adapt to deep learning models.

Ideally, a system could provide both functionalities: the high
performance of deep learning with the ability to interactively spot
issues and explore alternatives. However, the current architecture of
most neural networks makes it more challenging to examine
decisions of the model and locate problematic cases. Our work
tackles the following challenges and domain goals for seq2seq models
analogous to the three steps in rule-based systems:

\noindent \textbf{Goal G1 -- Examine Model Decisions:} It is first
important to examine the model's decision chain in order to track down
the error's root cause. As mentioned in
\autoref{sec:s2smodelsattention}, seq2seq models make decision through
several stages. While it has proven difficult to provide robust
examination in general-purpose neural networks, there has been success
for specific decision components.  For example, the attention stage
(S3) has proven specifically useful for inspection~\cite{xu2015show,bahdanau2014neural}.  Our first goal is to develop
interactive visual interfaces that help users understand the model's
components, their relationships, and pinpoint sources of error.

\noindent \textbf{Goal G2 -- Connect Decisions to Samples from
  Training Data:} Once a model makes a particular decision, a user
should be able to trace what factors influenced this decision.  While
it is difficult to provide specific reasoning about the many factors
that led to a decision in a trained model, we hope to provide other
means of analysis. In particular, we consider the approach of mapping
example states to those from previous runs of the model. For instance,
the training data defines the world view of a model and therefore
influences its learned decisions~\cite{koh2017understanding}. The goal
is to utilize (past) samples from training data as a proxy to better
understand the decision made on the example in question.

\noindent \textbf{Goal G3 -- Test Alternative Decisions:} Ultimately,
though, the goal of the user is to improve the model's performance and
robustness.  While the current state-of-the art for diagnosing and
improving deep neural network models is still in an early
stage~\cite{karpathy2015visualizing,smilkov2017direct,li2016understanding,
  kahng2018activis}, our goal is to allow the user to test
specific interventions. We aim to let the user investigate causal effects of changing parts of the model the let users ask \emph{what if} specific intermittent outputs of a model changed. 

Our motivating case study (\autoref{uc_dark}) follows these goals:
First, the user defines five hypotheses for causes of error and tests them by examining the model's decisions (G1).  Some of these decisions (for S1, S2) are represented in the model only as latent high-dimensional vectors. To make these parts tangible for the user, she connects them to representative neighbors from the training data (G2).
Finally, by probing alternatives in the beam search (G3) she finds a
temporary alternative that helps her to formulate a better solution.

We use these goals to compile a set of visualization and
interaction tasks for \texttt{Seq2Seq-Vis}. The mapping of these tasks
to goals is indicated by square brackets:

\noindent \textbf{Task T1 - Create common visual encodings} of all
five model stages to allow a user to examine the learned connections
between these modules. [G1]

\noindent \textbf{Task T2 - Visualize state progression} of latent
vector sequences over time to allow for high-level view of the learned
representations. [G1]

\noindent \textbf{Task T3 - Explore generated latent vectors and their
  nearest neighbors} by querying a large database of training examples
to facilitate error identification and training adjustment. [G2]

\noindent \textbf{Task T4 - Generate sensible alternative decisions
  for different stages of the model and compare them} to ease model
exploration and compare possible corrections. [G1, G3]

\noindent \textbf{Task T5 - Create a general and coherent interface}
to utilize a similar front-end for many sequence-to-sequence problems
such as translation, summary, and generation. [G1,G2,G3]

%\mib{I would love to have another pass here: e.g., common vis enc does or general interface does not fit wrt the level of detail}

In the following section, we will match these tasks and goals to design decisions for \tool.

\begin{figure*}[ht]
  \centering
  \includegraphics[width=\textwidth]{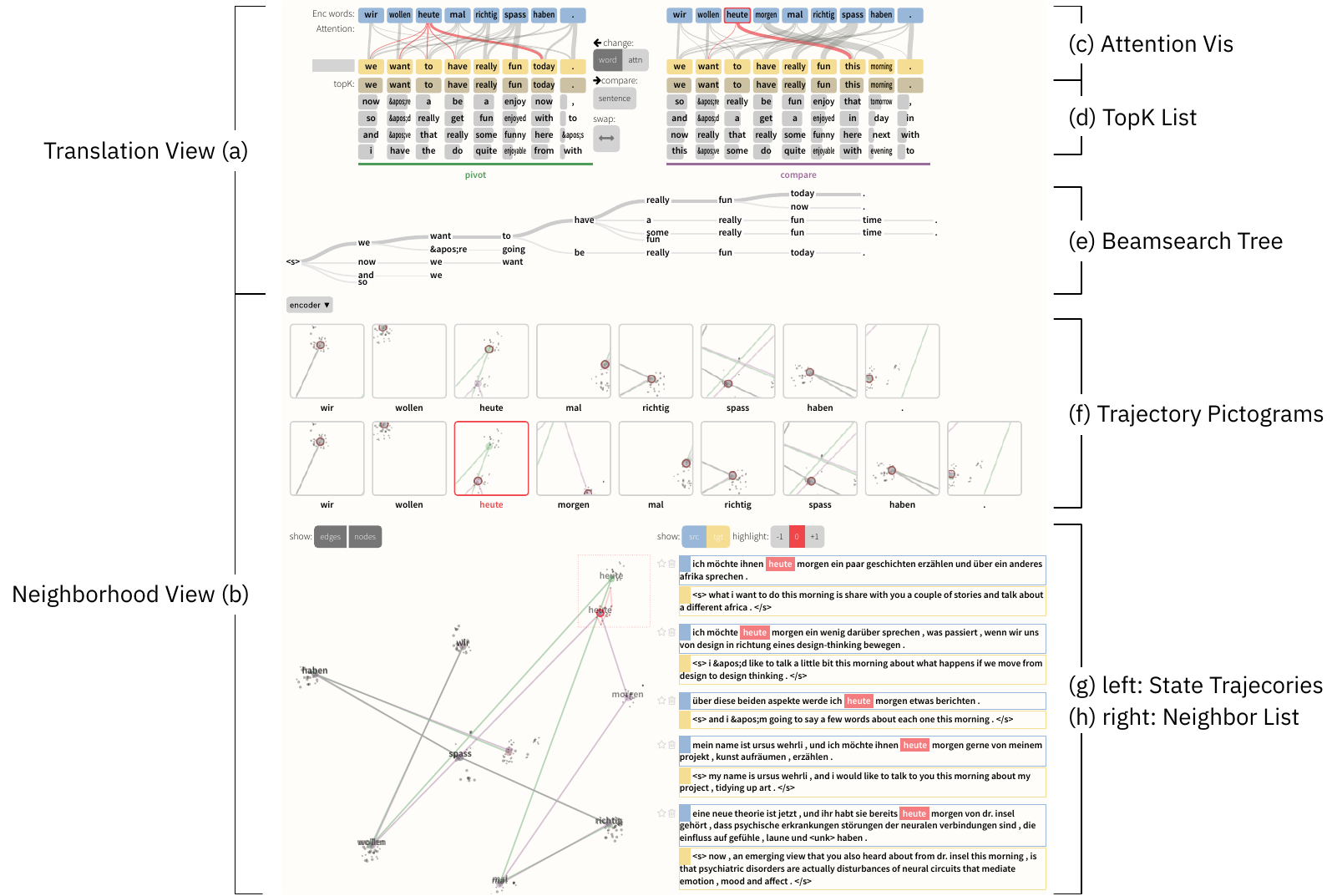}
  \caption{Overview of \texttt{Seq2Seq-Vis}: The two main views (a) Translation View and (b) Neighborhood View facilitate different modes of analysis. Translation View provides (c) visualizations for attention, (d) the top k word predictions for each time step, and (e) the beam search tree. The Neighborhood View goes deeper into what the model has learned by providing (f,g) a projection of state trajectories and (h) a list of nearest neighbors for a specific model state. }
	\label{fig:overview}

\end{figure*}

\section{Design of \texttt{Seq2Seq-Vis}}
\label{sec:design}

\texttt{Seq2Seq-Vis} is the result of an iterative design process and
discussions between experts in machine learning and visualization.  In
regular meetings we evaluated a series of low-fidelity prototypes
and tested them for usability. The design presented in this section
combines the prevailing ideas into a comprehensive tool.

\texttt{Seq2Seq-Vis} is composed of two main views facilitating
different modes of analysis: In the upper part, the
\textit{translation view} provides a visual encoding for each of the
model stages and fosters understanding and comparison tasks. In
the lower part, the \textit{neighborhood view} enables deep analysis
based on neighborhoods of training data. \autoref{fig:overview} shows
the complete tool.

\subsection{Translation View}
\label{sec:trans_view}

In the translation view (\autoref{fig:overview}a), each functional
stage of the seq2seq model is mapped to a visual encoding (T1, T2, G1).
We generalize and extend encodings from Olah \&
Carter~\cite{olah2016attention} and Le et al.~\cite{Le:2012td}.  In
Attention Vis (\autoref{fig:overview}c), the encoder words are shown
in blue, the decoder words in yellow, and the attention is shown
through weighted bipartite connections. To reduce visual clutter the
attention graph is pruned. For each decoder step all edges are
excluded that fall into the lower quartile of the attention
probability distribution.

Right below the yellow decoder words, the top $K$ predictions (S4 of model)
for each time step are shown (\autoref{fig:overview}d). Each possible
prediction encodes information about its probability in the
underlying bar chart, as well as an indication if it was chosen for the final output (yellow highlight).

In the bottom part of the translation view, a tree visualization shows
the hypotheses from the beam search
stage (\autoref{fig:overview}e). The most probable hypothesis, which
results in the final translation sentence, is highlighted.  Several
interactions can be triggered from the translation view, which will be
explained in \autoref{sec:interaction}.

\subsection{Neighborhood View}

The neighborhood view (\autoref{fig:overview}b) takes a novel approach
to look at model decisions in the context of finding similar examples
(T2, T3, G1, G2). As discussed in \autoref{sec:s2smodelsattention},
seq2seq models produce high-dimensional vectors at each stage, e.g.,
encoder states, decoder states, or context states.  It is difficult to
interpret these vectors directly. However, we can estimate their
meaning by looking at examples that produces similar vectors. To enable this
comparison, we precompute the hidden states of a large set of example
sentences (we use 50k sentences from the training set). For each state
produced by the model on a given example, \tool
searches for nearest neighbors from this large subset of precomputed
states.  These nearest neighbors are input to the \textit{state
  trajectories} (\autoref{fig:overview}g) and to the \textit{neighbor
  list} (\autoref{fig:overview}h).

The state trajectories show the changing internal hidden state of the
model with the goal of facilitating task T2.  This view encodes the
dynamics of a model as a continuous trajectory. First, the set for all
states and their closest neighbors are projected using a non-linear
algorithm, such as t-SNE \cite{maaten2008visualizing}, non-metric MDS
\cite{kruskal1964nonmetric}, or a custom projection (see
\autoref{sec:use_cases}). This gives a 2D positioning for each vector.
We use these positions to represent each encoder/decoder sequence as a
trace connecting its vectors. See \autoref{fig:overview}g for an
example of a trace representing the encoder states for \textit{wir
  wollen heute mal richtig spass haben}.

In the projection, the nearest neighbors to each vector are shown as
nearby dots.  When hovering over a vector from the input, the related
nearest neighbor counterparts are highlighted and a temporary red line
connects them. For vectors with many connections (high centrality), we
reduce visual clutter by computing a concave hull for all related
neighbors and highlight the related dots within the hull. Furthermore,
we set the radius of each neighbor dot to be dependent on how many
original states refer to it.  E.g., if three states from a decoder
sequence have one common neighbor, the neighbor's radius is set to
$\sim2.5$ (we use a $r(x) = \sqrt{2x}$ mapping with $x$ being number
of common neighbors).

%See \hs{FIGURE !} for an
%example.  \hs{@SG and @SR: should we talk about the linear model
%  experiments here ?}

The state trajectories can be quite long. To ease understanding,
we render a view showing states in their local neighborhood
as a series of trajectory pictograms (\autoref{fig:overview}f).
Each little window is a cut-out from the projection view, derived
from applying a regular grid on top of the projection plane.
Each pictogram shows only the cut-out region in which the
respective vector can be found.

Clicking on any projected vector will show the neighbor
list on the right side of the view. The neighbor list shows the actual
sentences corresponding to the neighbor points, grounding these
vectors in particular words and their contexts. Specifically, the
neighbor list shows all the nearest neighbors for the selected point
with the original sequence pair. The source or target position in the
sequence that matches is highlighted in red.  The user can facet the
list by filtering only to show source (blue) or target (yellow)
sequences.  She can also offset the text highlight by $-1$ or $+1$ to
see alignment for preceding or succeeding word positions (see
\autoref{fig:uc_dark_02}).

% From the projection view and the small multiples view the user can trigger a view on details about neighbors in the neighbor list.
% This list shows the nearest neighbor vectors within the sentences they were appearing in.
% For clarity, the user can hide all source or all target sentences.
% For several scenarios it is important to offset the highlights to the previous or succeeding word.
% The neighbor list should enable the user to see alignments within a state neighborhood of the training data.

\subsection{Global Encodings and Comparison Mode}

\texttt{Seq2Seq-Vis} uses visual encodings that are homogenous across
all views to create a coherent experience and ease the tool's
learning curve for model architects and trainers (T5). First, a
consistent color scheme allows the user to identify the stage and
origin of data (encoder - blue, decoder - yellow, pivot - green,
compare - violet).  Furthermore, every visual element with round
corners is clickable and leads to a specific action. Across all views,
hovering highlights related entities in red.

Additionally, the tool has a global comparison mode (T4, G1, G3).  As
soon as we generate a comparison sample from one of several triggers
(\autoref{sec:interaction}), all views switch to a mode that
allows comparison between examples.  Attention Vis, Trajectory
Pictograms, and State Projector display a superimposed
layer of visual marks labeled with a comparison color (violet)
different from the pivot color (green).  To ease understanding, we
disable all triggers in the comparison view.  However, by
providing an option to swap pivot and compare roles (arrow button), we
allow a consistent exploration flow from one sample to the next. The
only exception is the Beam Search Tree, which is only shown for the pivot
sample to save visual space.

\subsection{Interacting With Examples}
\label{sec:interaction}

% \texttt{Seq2Seq-Vis} allows multiple ways to interact with it's different parts to facilitate analysis of the model.
% Highlighting is propagated across all parts.

A major focus of \texttt{Seq2Seq-Vis} is interactive comparison
between different sources and targets (T4, G1, G3). We consider two
different modes of interactions to produce comparison samples or to
modify the pivot: \textbf{model-focused} and \textbf{language-focused}
changes.  Model-focused interactions let the user (model architect)
produce examples that the model believes are similar to the current
pivot to test small, reasonable variations for the different model
stages.  Language-focused interactions enable the user (model trainer)
to produce examples focussed on the language task and observe model
behavior.

%  in creating small
% variants of input or output to test model behavior, whilst
% language-focussed methods help observing the model when solving the
% translation task.

% Probing small changes for input or output means to modify high
% dimensional vectors.  A straight forward method would be changing the
% vector values directly. which comes with the cost of not being able to
% interpret these vectors.

For the model-focused interactions, we utilize a different variant of
neighborhoods. To replace a word with a slightly different, but
interpretable substitute, we search for neighbors from the model's
word vectors.  The user can trigger the substitution process by
clicking on the word to be replaced.  As a result, a word cloud
projecting the closest words w.r.t. their vector embedding in a 2D
plane is shown. A click on one of the words in the cloud replaces the
original and triggers a new translation in comparison mode.

\begin{figure}[ht]
\centering
\includegraphics[width=\columnwidth]{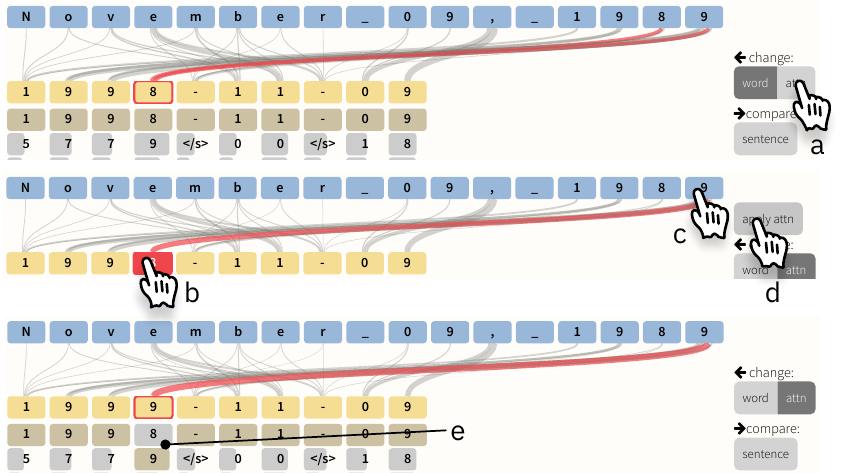}
\caption{To re-direct attention in \texttt{Seq2Seq-Vis}, the user first observes a split of attention between the input \textit{8} and \textit{9} for converting the last digits of a year in a date conversion model. She can  (a) select attention mode, (b) select the decoder word, (c) click on the preferred encoder word, (d) apply the attention change, and (e) see the models reaction. }
\label{fig:attn_interact}
\end{figure}

Another model-focused interaction is to modify the model directly,
for instance, by altering attention weights (S3 in model).
For this step, the user can switch to attention
modification and select a target word
for which attention should be modified. By repeatedly clicking on
encoder words, she gives more weights to these encoder
words. \autoref{fig:attn_interact} shows how the attention can be modified for an example. After hitting \textit{apply attn}, the attention is applied
for this position, overwriting the original attention distribution.

For language-focused interactions, the user can specify direct changes
to either the source or the target.  The user can trigger the changes
by using the \textit{manual compare} button and enter a new source or
a new target sentence. When the source is changed, a new full translation
is triggered. If the target is changed, a prefix decode is triggered
that constrains the search on a predefined path along the words
entered and continues regular beam search beyond.

Alternatively, the user can select the word from the top $K$ predictions
(\autoref{fig:overview}d) that seems to be the best
next word. By clicking on one of these words, a
prefix decode is triggered as described above and shown in
\autoref{fig:uc_dark_04}.

Initiating either of these interactions switches \texttt{Seq2Seq-Vis}
into comparison mode.  As a core analysis method, comparison allows to
derive insights about model mechanics (model-focused probing) or how
well the model solves the task (language-focused testing).

\subsection{Design Iterations}

\begin{figure}
\centering
\includegraphics[width=\columnwidth]{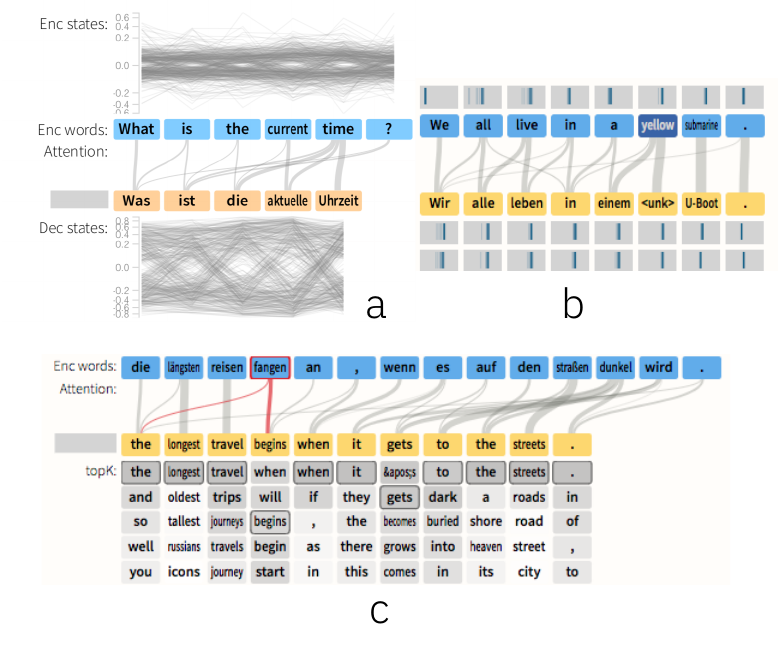}
\caption{Design variants for additional token information: (a) progression of hidden states, (b) density of neighborhoods, or (c) top $K$ predictions as heatmap.}
\label{fig:attention_vars}
\end{figure}

We considered several different variants for both main views of the system.
For the translation view, we considered incorporating more state
information directly into the encoding.  \autoref{fig:attention_vars}
shows iterations for adding per-word information around encoder and
decoder.  Similar to LSTMVis~\cite{strobelt2018lstmvis}, the hidden
state line charts show progression along encoder and decoder hidden
states (\autoref{fig:attention_vars}a). Domain scientists rejected this as too noisy for the
given domain goals. In a later iteration the visualization experts
proposed to indicate the closeness of the nearest neighbors with a
simple histogram-like encoding (\autoref{fig:attention_vars}b).
This information did not help to
formulate hypotheses. In addition, it did not reveal a lot of variance
(see abundance of similar small gray boxes).  The next
design focused on incorporating language features rather than latent
vectors. It showed for each time step of the decoder the top $K$
predicted words being produced as if there was only the top beam
evaluated until then.  Finally, we decided to
use the stronger visual variable \textit{length} to encode the
probability values (\autoref{fig:overview}d).

In the neighborhood view, the trajectory pictograms are a result of a
series of visual iterations around combining the linear nature of
sequences with keeping some spatial information describing vector
proximity. We divide the state trajectory view into cutout regions forming a regular grid.
Using a regular grid limits the variants
of basic pictograms to a small and recognizable number.  Alternative
ideas were to center the cutout area around each state or
to use the bounding box of the sequence state and all its neighbors as
area. Both alternatives created highly variant
multiples that introduced visual noise.
For the regular grid, choosing the right grid size is important and
the current static solution of applying a $3x3$ grid will be replaced 
by a non-linear function of number of displayed words to allow
for scalability.

%The regular grid has the drawback that
%states can land close to the border of these segments, but its largest
%advantage is that it keeps the variety of unique cut-outs small and
%allows recognition of the same regions.
% Alternatives cutout
% ideas would be to center the cutout area around each state or to use
% the bounding box of the sequence state and all its neighbors as area.
% Both alternatives would create unique and highly variant multiples for
% each state.

%\subsection{Selection and Discovery}
%\label{sec:select}
%
%\subsection{Compare and Diagnose}
%\label{sec:compare}

\section{Implementation}
\label{sec:implementation}

\texttt{Seq2Seq-Vis} allows for querying and interaction with a live
system. To facilitate this, it uses tight integration of a seq2seq
model with the visual client. We based the interface between
both parts on a REST API, and we used
OpenNMT~\cite{2017opennmt} for the underlying model framework.
We extended the core OpenNMT-py distribution to allow easy access to
latent vectors, the search beams, and the attention values.
Furthermore, we added non-trivial model-diagnostic modifications 
for translation requests to allow prefix decoding and to apply
user-specific attention. We plan to distribute \texttt{Seq2Seq-Vis} as
the default visualization mode for OpenNMT.

To allow fast nearest neighbor searches, Python scripts extract
the hidden state and context values from the model for points in a
large subset of the training data.  These states are saved in HDF5
files and indexed utilizing the Faiss~\cite{JDH17} library to allow
fast lookups for closest dot products between vectors.  For TSNE and
MDS projections we use the SciKit Learn package~\cite{scikit-learn}
for Python.

The model framework and the index work within a Python Flask server to
deliver content via a REST interface to the client. The client is
written in Typescript. Most visualization components are using the
d3js library. Source code, a demo instance, and a descriptive webpage are available at \url{http://seq2seq-vis.io}.

\section{Use Cases}
\label{sec:use_cases}

We demonstrate the application of \texttt{Seq2Seq-Vis} and how it
helps to generate insights using examples from a toy date conversion
problem, abstractive summarization, and machine translation
(\autoref{uc_dark}).

\noindent \textbf{Date Conversion.} Seq2seq models can be
difficult to build and debug even for simple problems.  A common test
case used to check whether a model is implemented correctly is to
learn a well-specified deterministic task.  Here we
consider the use case of converting various date formats
to the unified format YEAR-MONTH-DAY. For example, the source
\textit{March 25, 2000} should be converted to the
target \textit{2000-03-25}.  While this problem is much
simpler than language translation, it tests the different components
of the system. Specifically, the encoder (S1) must learn to identify
different months, the attention (S3) must learn to reorder between the
source and the target, and the decoder (S2) must express the source
word in a numeric format.

\begin{figure}[t]
  \centering
%  \includegraphics[width=\linewidth]{Screenshots/ScreeshotDates2}
%  \vspace{0.4cm}
%
%  \includegraphics[height=5cm]{Screenshots/ScreenPath}
	\includegraphics[]{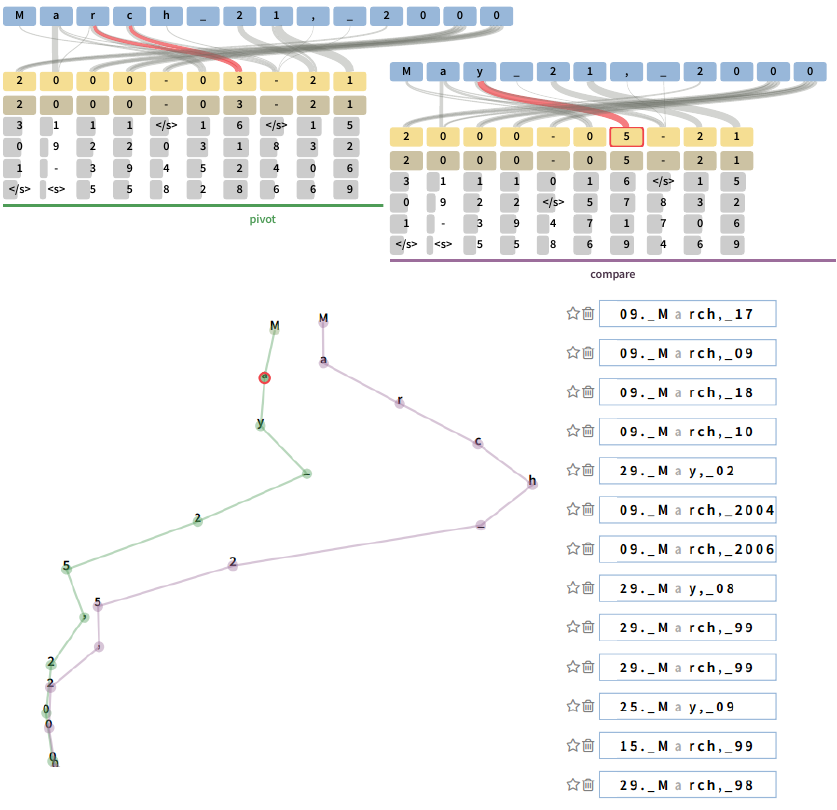}
  \caption{Comparing translations for a date conversion model. The input sequences \textit{March 21, 2000} and \textit{May 21, 2000} are only different by some letters. The attention (top) for predicting the correct months 3 and 5 is focused on this difference (\textit{y} vs. \texttt{rc}). The trajectory view (bottom left) shows this difference along the progression of encoder states. The neighborhood list (bottom right) indicates that after input of \textit{M a} the model is still undecided.}
\label{fig:dates}
\end{figure}

\tool provides tools for examining these different stages of the
model. Figure~\ref{fig:dates} shows an example, where the
user, following Goal 3, employs a comparison between two different
translations, one starting with \textit{March} and the other with
\textit{May}.  These two translations are nearly identical, except one
yields the month \textit{3} and the other \textit{5}. Following Goal
1, the user might want to examine the models decisions. The upper
translation view provides a way to compare between the attention on
the two inputs. The red highlighted connections indicate that the
first sentence attention focuses on \textit{r c} wheres the second focuses on \textit{y}. These characters are used by the model to distinguish the two
months since it cannot use \textit{M a}. The user can also
observe how the encoder learns to use these letters. The trajectory
view compares the encoder states of sentence 1 and sentence 2. Here we
use a custom projection, where the y-axis is the relative position of a word in a sentence and the
x-axis is a 1-d projection of the vector. This reveals that the
two trajectories are similar before and after these characters, but
diverge significantly around \textit{r} and \textit{c}. Finally,
following Goal 2, the user can connect these decisions back to the training
data. On the right, she can see the nearest neighbors around the
letter \textit{a} in \textit{M a y} (highlighted in
red). Interestingly, the set of nearest neighbors is almost equally
split between examples of \textit{M a y} and \textit{M a r c h},
indicating that at this stage of decoding the model is preserving
its uncertainty between the two months.

% A model needs to learn how to reo
% Due to the
% different orders in which the three information can appear, the model
% has to learn to identify the type of each input and align it to the
% appropriate location in the generated output. We consider each
% individual character as separate input to the model and generate the
% new format character by character.

\noindent\textbf{Abstractive Summarization.} For our
second use case we apply the tool to a summarization
problem. Recently, researchers have developed methods for
\textit{abstractive} text summarization that learn how to produce a
shorter summarized version of a text passage. Seq2seq models are
commonly used in this framework~\cite{rush2015neural,nallapati2016abstractive,paulus2017deep,see2017get}. In abstractive summarization, the target passage may not contain the same phrasing as the original. Instead, the model learns to paraphrase and alter the wording in the process of summarization.

Studying how paraphrasing happens in seq2seq systems is a core
research question in this area. Rush et al.~\cite{rush2015neural} describe a system using the Gigaword data set (3.8M sentences). They study the example
source sentence \textit{russian defense minister ivanov called sunday
  for the creation of a joint front for combating global terrorism} to
produce a summary \textit{russia calls for joint front against
  terrorism}.  Here \textit{russia} compresses the phrase
\textit{russian defense minister ivanov} and \textit{against}
paraphrases \textit{for combating}.

To replicate this use case we consider a user analyzing this sentence.
In particular, he is interested in understanding how the model selects
the length and the level of abstraction. He can analyze
this in the context of Goal 3, testing alternatives predictions of the model, in particular targeting Stage 4. As
discussed in Sect~\ref{sec:design}, \tool shows the top $K$ predictions at
each time step. When the user clicks on a prediction, the system will
produce a sentence that incorporates this prediction. Each choice is
``locked'' so that further alterations can be made.

\begin{figure}
  \centering
  \includegraphics[width=\linewidth]{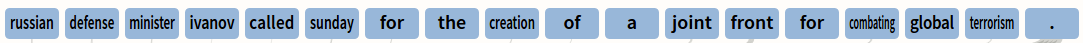}
  \vspace{0.01cm}

  \includegraphics[height=2.4cm]{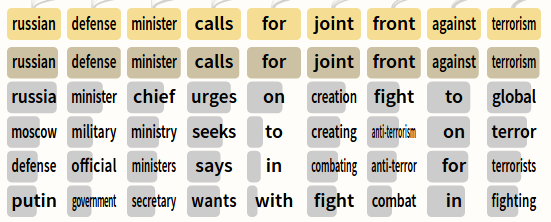}
  \vspace{0.2cm}

  \includegraphics[height=2.4cm]{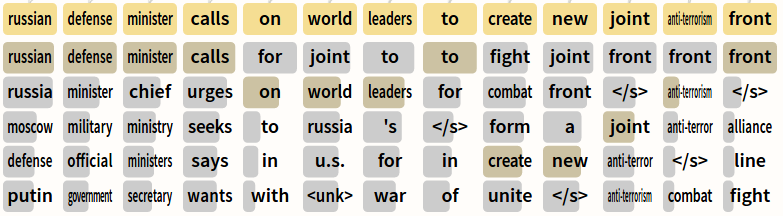}
  \vspace{0.2cm}

  \includegraphics[height=2.4cm]{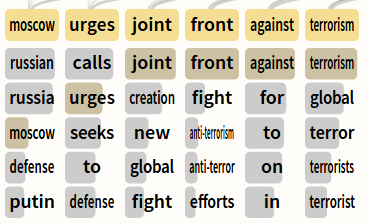}\hspace{0.3cm}
  \includegraphics[height=2.4cm]{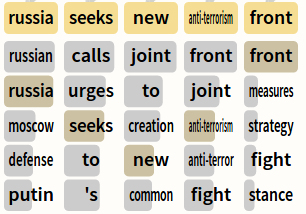}
  \caption{Use case of abstractive summarization. The input sentence \textit{russian defense minister ivanov called sunday for the creation of a joint front for combating global terrorism} can be summarized in different ways. The yellow boxes indicate alternative translations for different prefix decode settings. Top: the unconstrained abstraction; middle: changing prediction from \textit{for} to \textit{on} leads to automatic insertion of \textit{on world leaders} to stay grammatically correct; bottom left: changing the first word from \textit{russian} to \textit{moscow} or \textit{russia} compresses the sentence even more while retaining its meaning.}
  \label{fig:abssum}
\end{figure}

\autoref{fig:abssum} shows the source input to this model. We
can see four different summarizations that the model produces based on
different word choices. Interestingly, specific local choices do have
a significant impact on length, ranging from five to thirteen
words. 
%Furthermore, the model reacts to word choices by altering its path.
Switching from \textit{for} to \textit{on} leads the decoder to
insert an additional phrase \textit{on world leaders} to maintain
grammaticality. While the model outputs the top choice, all other choices have relatively high probabilities. This observation has motivated research into adding constraints to the prediction at each time step. Consequently, we have added methods for constraining length and prediction into
the underlying seq2seq system to produce different outputs.

% Another common task is abstractive summarization. Here, $x$ is a long
% sequence and $y$ a short sequence in the same language. The task is to
% compress and paraphrase the input. We use the example of news headline
% generation using the Gigaword dataset~\sg{[?]}. One example has the
% input ``russian defense minister ivanov called sunday for the creation
% of a joint front for combating global terrorism'' and the output
% ``russia calls for joint front against terrorism''. Since the model
% has to handle many previously unseen words such as names of entities,
% a common error is that they generate the wrong entity at a position.

\noindent \textbf{Machine Translation.}
Finally, we consider a more in-depth use case of a real-world
machine translation system using a complete model trained
on WMT '14 (3.96M examples) to translate from German to English.
This use case considers a holistic view of how an expert might go about
understanding the decisions of the system.

\begin{figure}
  \centering
%  \includegraphics[width=\linewidth]{Screenshots/spoke3}
%  \vspace{0.2cm}
%
%  \includegraphics[width=\linewidth]{Screenshots/spoke2}
\includegraphics{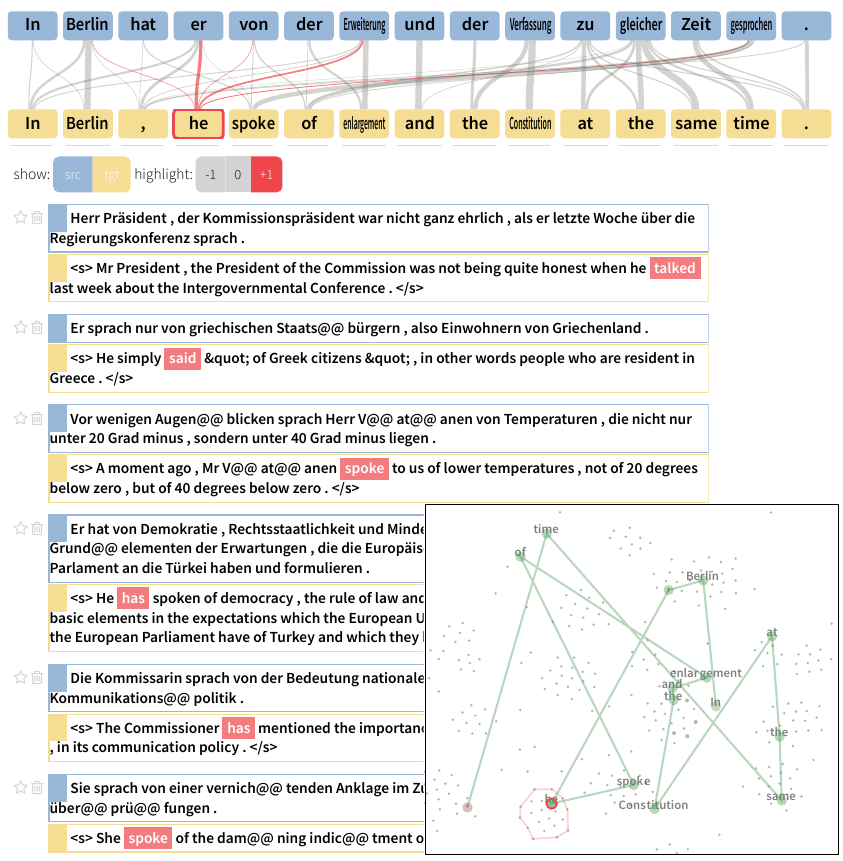}
  \caption{Use case language translation using WMT'14 data. The attention graph (top) shows how attention for the target word \textit{he} is not only focused on the decoder counterpart \textit{er} but also on the following words, even to the far away verb \textit{gesprochen} (spoke). The state trajectory (bottom left) for the decoder states reveals how close \textit{he} and \textit{spoke} are. The neighborhood list indicates that the model sets the stage for predicting \textit{spoke} as next word.}
    \label{fig:spoke}
\end{figure}

Figure~\ref{fig:spoke} shows an example source input and its
translation. Here the user has input a source sentence, translated it,
and activated the neighbor view to consider the decoder states. She is
interested in better understanding each stage of the model at this
point. This sentence is interesting as there is significant reordering
that must occur to translate from the original German to English. For
instance, the subject \textit{he} is at the beginning of the clause, but must interact with the verb \textit{gesprochen} at the end of the German sentence.

We consider Goals 1 and 2 applied to this example, with the
intent of analyzing the encoder, decoder, attention, and prediction (S1-S4). First we look at the attention. Normally, this stage focuses on
the word it is translating (\textit{er}), but researchers have noted
that neural models often look ahead to the next word in this process~\cite{koehn2017six}. We can see branches going from \textit{he}
to potential next steps (e.g., \textit{von} or \textit{gesprochen}). We
can further view this process in the decoder trajectory shown below,
where \textit{he} and \textit{spoke} are placed near each other in the
path. Hovering over the vector \textit{he} highlights it globally in
the tool. Furthermore, if we click on \textit{he}, we can link this
state to other examples in our data (Goal 2). On the right we can see these
related examples, with the next word (+1) highlighted.  We find that
the decoder is representing not just the information for the current
word, but also anticipating the translation of the verb
\textit{sprechen} in various forms.

\begin{figure*}
  \centering
  \includegraphics[width=\textwidth]{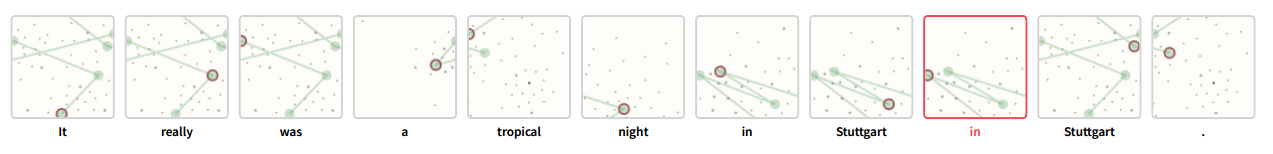}
  \caption{An under-trained English-German model. Repetition repetition is a commonly observed phenomenon in under-trained or under-parametrized models. Here the trajectory pictograms show that for the repetition \textit{in Stuttgart in Stuttgart} the decoder states alternate in the same region before being able to break apart.}
    \label{fig:repeat}
\end{figure*}

In this case we are seeing the model behaving correctly to produce a
good translation. However, the tool can also be useful when there are
issues with the system. One common issue in under-trained or
under-parameterized seq2seq models is to repeatedly generate the same
phrase. Figure~\ref{fig:repeat} shows an example of this
happening. The model repeats the phrase \textit{in Stuttgart in
  Stuttgart}. We can easily see in the pictogram view that the
decoder model has produced a loop, ending up in nearly the same
position even after seeing the next word. As a short-term fix, the
tool's prefix decoding can get around this issue. It remains an
interesting research question to prevent this type of cycle from occurring in general.

% As mentioned before, the most common application for
% \emph{sequence-to-sequence} models is machine translation. Here, $x$
% is a sentence in one language and $y$ the same sentence in another
% language.  These models heavily depend on the availability of training
% data~\sg{[?]}. Thus, we show one model with limited data and one
% trained on large data.  Specifically, we show models for the IWLST '14
% (160k examples)~\sg{[?]} and WMT '14 (3.96M examples)~\sg{[?]}
% English-German datasets.  While most relevant, understanding the
% output of the model requires fluency of both languages in the model.
% =======
% \noindent \textbf{Common Machine Translation Issues}
% As mentioned before, the most common application for
% \emph{sequence-to-sequence} models is machine translation. Here, $x$
% is a sentence in one language and $y$ the same sentence in another
% language.  These models heavily depend on the availability of training
% data~\cite{sennrich2015improving}. Thus, we show one model with limited data and one
% trained on large data.  Specifically, we show models for the IWLST '14
% (160k examples)~\cite{mauro2012wit3} and WMT '14 (3.96M examples)~\cite{bojar2014wmt}
% English-German datasets.  While most relevant, understanding the
% output of the model requires fluency of both languages in the model.
% >>>>>>> 3d034fbd67078d5bb1d89969354b85d4d25e1d63

\section{Related Work}
\label{sec:related-work}

%It is a difficult task to understand explanations that are generated for deep learning model predictions that are generated by various methods~\cite{biran2017explanation, narayanan2018humans}. 
Various methods~\cite{biran2017explanation,narayanan2018humans} have been proposed to generate explanations for deep learning model predictions. Understanding them still remains a difficult task.
To better address the specific issues of our users, we narrow the target audience for our proposed tool. Following the classifications by Strobelt et al.~\cite{strobelt2018lstmvis} and Hohman et al.~\cite{hohman2018visual}, our tool aims at model developers who have at least a conceptual understanding of how the model works. This is opposed to end users, who are agnostic to the technique used to arrive at a specific result. 
Following Hohman et al., analysis itself can broadly be divided into \emph{global model analysis} and \emph{instance-based analysis}. In global model analysis, the most commonly seen methods are visualizations of the internal structure of trained deep learning models. Instance-based analysis may be coupled with interactive experimentation with the goal of understanding a particular prediction using the local information around only one input~\cite{park2018conceptvector}.

\paragraph{Global Model Analysis} 
Most recent work focuses on visualizing hidden representations of convolutional neural networks (CNNs)~\cite{Lecun:1998hy} for computer vision applications.
Techniques for visualizing CNNs include showing neural activity in the convolutional layers as overlay over the image~\cite{Erhan:2009wa, Le:2012td} and directly showing the images that maximize the activity~\cite{simonyan2013deep}. Zeiler and Fergus~\cite{zeiler2014visualizing} use
deconvolutional networks to explore the layers of a CNN. This approach is widely used to generate explanations of models, for example, by Yosinski et al.~\cite{yosinski2015understanding}.

A similar line of work has focused on visualizing recurrent neural networks (RNNs) and other sequence models. Preliminary work by Karpathy et al.~\cite{karpathy2015visualizing} uses static visualizations to understand hidden states in language models. They demonstrate that selected cells can model clear events such as open parentheses and the start of URLs. Strobelt et al.~\cite{strobelt2018lstmvis} introduce LSTMVis, an interactive tool that allows users to understand activation patterns of combinations of hidden states. LSTMVis shows the neighborhood of activations within the training data as an approach of making sense of the complex interactions in a context. 

Similar to our approach, Kahng et al.~\cite{kahng2018activis} propose using the model structure as the entry point into the analysis. In their approach, they try to understand connections between misclassified examples and hidden states of parts of the network by showing activation pattern differences between correct and false examples from the training data. 
Ming et al.~\cite{ming2017understanding} propose RNNVis, a tool that uses word clouds instead of full contexts or sentences to show typical words that appear for activation patterns. 
Our approach to show embeddings of a whole phrase is similar to that of Johnson et al.~\cite{johnson2016google}. They use three-dimensional tSNE in order to visualize progressions of context vectors. Novel in our approach are different types of progressions as well as the connection and embedding with neighborhoods. 

An alternative to visualizing what a model has learned is visualizing how it is learning. RNNbow by Cashman et al.~\cite{cashman2017rnnbow} shows the gradient flow during backpropagation training in RNNs to visualize how the network is learning.

\paragraph{Instance-Based Analysis}

Instance-based analysis is commonly used to understand local decision boundaries and relevant features for a particular input.
For example, Olah et al.~\cite{olah2018building} extend methods that compute activations for image classification to build an interactive system that assesses specific images. They show that not only the learned filters of a CNN matter, but also their magnitudes. 
The same type of analysis can be used to answer counter-factual ``what if'' questions to understand the robustness of a model to pertubations. Nguyen et al.~\cite{nguyen2015deep} show that small perturbations to inputs of an image classifier can drastically change the output. Interactive visualization tools such as Picasso~\cite{henderson2017picasso} can manipulate and occlude parts of an image as input to an image classifier.
Krause et al.~\cite{krause2016interacting} use partial dependence diagnostics to explain how features affect the global predictions, while users can interactively tweak feature values and see how the prediction responds to instances of interest.

There is an intrinsic difficulty in perturbing inputs of models that operate on text. While adding noise to an image can be achieved by manipulating the continuous pixel values, noise for categorical text is less well defined. However, there is a rich literature for methods that compute relevant inputs for specific predictions, for example by computing local decision boundaries or using gradient-based saliency~\cite{ribeiro2016should, ross2017right, Zintgraf17visualizing, Alvarez17causal, li2015visualizing}.
Most of these methods focus on classification problems in which only one output exists.  Ruckle et al.~\cite{ruckle2017end} address this issue and extend saliency methods to work with multiple outputs in a question-answering system. As an alternative to saliency-methods, Ding et
al.~\cite{ding2017visualizing} use a layer-wise relevance propagation technique~\cite{bach2015pixel} to understand relevance of input with regard to an output in sequence-to-sentence models. 
Yet another approach to understand predictions within text-based models is to find the minimum input that still yields the same prediction~\cite{Lei2016rationalizing,li2016understanding}. 
None of the previous methods use our approach of using nearest neighbors of word embeddings to compare small perturbations of RNNs.

One commonality among all these approaches is that they treat the model as a black box that generates a prediction. In contrast, we are assuming that our users have an understanding of the different parts of a sequence-to-sequence model. Therefore, we can use more in-depth analysis, such as interactive manipulations of input, output, and attention. Our beam search and attention manipulations follow the approach by Lee et al.~\cite{lee2017interactive} who show a basic prototype to manipulate these parts of a model.

%In particular, we believe that methods from global analysis can help with this task. Therefore, we combine methods from both analysis types to (1) operate on user-input to encourage experimentation, (2) compare two specific instances, (3) use on projections of trajectories of sequences, (4) extract information from the training data to assist with the analysis. 

\section{Conclusions and Future Work}\label{sec:conclusion}

\texttt{Seq2Seq-Vis} is a tool to facilitate deep exploration of
all stages of a seq2seq model. We apply our set of goals to deep learning models that are traditionally difficult to interpret. 
To our knowledge, our tool is the first of its kind to combine insights about model mechanics (translation view) with insights about model semantics (neighborhood view), while allowing for "what if"-style counterfactual changes of the
model's internals. 

Being an open source project, we see future work in evaluating the longitudinal feedback from real-world users for suggested improvements. Two months after release, we already observed some initial quantitative and qualitative feedback. Currently, more then 5,500 page views have been recorded and 156 users liked (starred) the project on Github. The most requested new feature is integration of the tool with other ML frameworks.

There are many avenues for future work on the algorithmic and visualization side. 
Improving the projection techniques to better respect the linear order of sequences would be helpful. 
The tool could be extended to different sequence types, including audio, images, and video. 
Supporting these different data types requires non-trivial expansion of visual encoding for input and output.
A prerequisite for future work targeting different models and frameworks is that model architects implement open models with hooks for observation \textit{and} modification of model internals.  
We hope that \tool will inspire novel visual and algorithmic methods to fix models without retraining them entirely. 
%\comment{add more if possible}

%% if specified like this the section will be committed in review mode
%\acknowledgments{
%The authors wish to thank A, B, and C. This work was supported in part by
%a grant from XYZ (\# 12345-67890).}

%\bibliographystyle{abbrv}
\bibliographystyle{abbrv-doi}

\bibliography{template}

\begin{thebibliography}{10}

\bibitem{Alvarez17causal}
D.~Alvarez{-}Melis and T.~S. Jaakkola.
\newblock A causal framework for explaining the predictions of black-box
  sequence-to-sequence models.
\newblock In {\em Proceedings of the 2017 Conference on Empirical Methods in
  Natural Language Processing, {EMNLP} 2017, Copenhagen, Denmark, September
  9-11, 2017}, pp. 412--421, 2017.

\bibitem{bach2015pixel}
S.~Bach, A.~Binder, G.~Montavon, F.~Klauschen, K.-R. M{\"u}ller, and W.~Samek.
\newblock On pixel-wise explanations for non-linear classifier decisions by
  layer-wise relevance propagation.
\newblock {\em PloS one}, 10(7):e0130140, 2015.

\bibitem{bahdanau2014neural}
D.~Bahdanau, K.~Cho, and Y.~Bengio.
\newblock Neural machine translation by jointly learning to align and
  translate.
\newblock {\em arXiv preprint arXiv:1409.0473}, 2014.

\bibitem{biran2017explanation}
O.~Biran and C.~Cotton.
\newblock Explanation and justification in machine learning: A survey.
\newblock In {\em IJCAI-17 Workshop on Explainable AI (XAI)}, p.~8, 2017.

\bibitem{cashman2017rnnbow}
D.~Cashman, G.~Patterson, A.~Mosca, and R.~Chang.
\newblock Rnnbow: Visualizing learning via backpropagation gradients in
  recurrent neural networks.
\newblock In {\em Workshop on Visual Analytics for Deep Learning (VADL)}, 2017.

\bibitem{ding2017visualizing}
Y.~Ding, Y.~Liu, H.~Luan, and M.~Sun.
\newblock Visualizing and understanding neural machine translation.
\newblock In {\em Proceedings of the 55th Annual Meeting of the Association for
  Computational Linguistics (Volume 1: Long Papers)}, vol.~1, pp. 1150--1159,
  2017.

\bibitem{Erhan:2009wa}
D.~Erhan, Y.~Bengio, A.~Courville, and P.~Vincent.
\newblock {Visualizing higher-layer features of a deep network}.
\newblock Technical report, University of Montreal, 2009.

\bibitem{gehring2016convenc}
J.~Gehring, M.~Auli, D.~Grangier, and Y.~N. Dauphin.
\newblock {A Convolutional Encoder Model for Neural Machine Translation}.
\newblock {\em ArXiv e-prints}, Nov. 2016.

\bibitem{gehring2017convs2s}
J.~Gehring, M.~Auli, D.~Grangier, D.~Yarats, and Y.~N. Dauphin.
\newblock {Convolutional Sequence to Sequence Learning}.
\newblock {\em ArXiv e-prints}, May 2017.

\bibitem{awadalla2018achieving}
H.~Hassan~Awadalla, A.~Aue, C.~Chen, V.~Chowdhary, J.~Clark, C.~Federmann,
  X.~Huang, M.~Junczys-Dowmunt, W.~Lewis, M.~Li, S.~Liu, T.-Y. Liu, R.~Luo,
  A.~Menezes, T.~Qin, F.~Seide, X.~Tan, F.~Tian, L.~Wu, S.~Wu, Y.~Xia,
  D.~Zhang, Z.~Zhang, and M.~Zhou.
\newblock Achieving human parity on automatic chinese to english news
  translation.
\newblock March 2018.

\bibitem{henderson2017picasso}
R.~Henderson and R.~Rothe.
\newblock Picasso: A modular framework for visualizing the learning process of
  neural network image classifiers.
\newblock {\em Journal of Open Research Software}, 5(1), 2017.

\bibitem{hern2017}
A.~Hern.
\newblock Facebook translates 'good morning' into 'attack them', leading to
  arrest.
\newblock {\em The Guardian}, Oct 2017.

\bibitem{hohman2018visual}
F.~Hohman, M.~Kahng, R.~Pienta, and D.~H. Chau.
\newblock Visual analytics in deep learning: An interrogative survey for the
  next frontiers.
\newblock {\em arXiv preprint arXiv:1801.06889}, 2018.

\bibitem{JDH17}
J.~Johnson, M.~Douze, and H.~J{\'e}gou.
\newblock Billion-scale similarity search with gpus.
\newblock {\em arXiv preprint arXiv:1702.08734}, 2017.

\bibitem{johnson2016google}
M.~Johnson, M.~Schuster, Q.~V. Le, M.~Krikun, Y.~Wu, Z.~Chen, N.~Thorat,
  F.~Vi{\'e}gas, M.~Wattenberg, G.~Corrado, et~al.
\newblock Google's multilingual neural machine translation system: enabling
  zero-shot translation.
\newblock {\em arXiv preprint arXiv:1611.04558}, 2016.

\bibitem{kahng2018activis}
M.~Kahng, P.~Y. Andrews, A.~Kalro, and D.~H.~P. Chau.
\newblock Activis: Visual exploration of industry-scale deep neural network
  models.
\newblock {\em IEEE transactions on visualization and computer graphics},
  24(1):88--97, 2018.

\bibitem{karpathy2015visualizing}
A.~Karpathy, J.~Johnson, and F.-F. Li.
\newblock {Visualizing and understanding recurrent networks}.
\newblock {\em ICLR Workshops}, 2015.

\bibitem{2017opennmt}
G.~{Klein}, Y.~{Kim}, Y.~{Deng}, J.~{Senellart}, and A.~M. {Rush}.
\newblock {OpenNMT: Open-Source Toolkit for Neural Machine Translation}.
\newblock {\em ArXiv e-prints}.

\bibitem{koehn2017six}
P.~Koehn and R.~Knowles.
\newblock Six challenges for neural machine translation.
\newblock {\em arXiv preprint arXiv:1706.03872}, 2017.

\bibitem{koh2017understanding}
P.~W. Koh and P.~Liang.
\newblock Understanding black-box predictions via influence functions.
\newblock {\em arXiv preprint arXiv:1703.04730}, 2017.

\bibitem{krause2016interacting}
J.~Krause, A.~Perer, and K.~Ng.
\newblock Interacting with predictions: Visual inspection of black-box machine
  learning models.
\newblock In {\em Proceedings of the 2016 CHI Conference on Human Factors in
  Computing Systems}, pp. 5686--5697. ACM, 2016.

\bibitem{kruskal1964nonmetric}
J.~B. Kruskal.
\newblock Nonmetric multidimensional scaling: a numerical method.
\newblock {\em Psychometrika}, 29(2):115--129, 1964.

\bibitem{Le:2012td}
Q.~V. Le, M.~Ranzato, R.~Monga, M.~Devin, G.~Corrado, K.~C. 0010, J.~Dean, and
  A.~Y. Ng.
\newblock {Building high-level features using large scale unsupervised
  learning.}
\newblock {\em ICML}, 2012.

\bibitem{Lecun:1998hy}
Y.~Lecun, L.~Bottou, Y.~Bengio, and P.~Haffner.
\newblock {Gradient-based learning applied to document recognition}.
\newblock {\em Proceedings of the IEEE}, 86(11):2278--2324, 1998.

\bibitem{lee2017interactive}
J.~Lee, J.-H. Shin, and J.-S. Kim.
\newblock Interactive visualization and manipulation of attention-based neural
  machine translation.
\newblock In {\em Proceedings of the 2017 Conference on Empirical Methods in
  Natural Language Processing: System Demonstrations}, pp. 121--126, 2017.

\bibitem{Lei2016rationalizing}
T.~Lei, R.~Barzilay, and T.~S. Jaakkola.
\newblock Rationalizing neural predictions.
\newblock In {\em Proceedings of the 2016 Conference on Empirical Methods in
  Natural Language Processing, {EMNLP} 2016, Austin, Texas, USA, November 1-4,
  2016}, pp. 107--117, 2016.

\bibitem{li2015visualizing}
J.~Li, X.~Chen, E.~Hovy, and D.~Jurafsky.
\newblock {Visualizing and Understanding Neural Models in NLP}.
\newblock In {\em NAACL}, pp. 1--10. Association for Computational Linguistics,
  San Diego, California, jun 2016.

\bibitem{li2016understanding}
J.~Li, W.~Monroe, and D.~Jurafsky.
\newblock Understanding neural networks through representation erasure.
\newblock {\em arXiv preprint arXiv:1612.08220}, 2016.

\bibitem{liu2018generating}
P.~J. Liu, M.~Saleh, E.~Pot, B.~Goodrich, R.~Sepassi, L.~Kaiser, and
  N.~Shazeer.
\newblock Generating wikipedia by summarizing long sequences.
\newblock {\em arXiv preprint arXiv:1801.10198}, 2018.

\bibitem{maaten2008visualizing}
L.~v.~d. Maaten and G.~Hinton.
\newblock Visualizing data using t-sne.
\newblock {\em Journal of machine learning research}, 9(Nov):2579--2605, 2008.

\bibitem{mauro2012wit3}
C.~Mauro, G.~Christian, and F.~Marcello.
\newblock Wit3: Web inventory of transcribed and translated talks.
\newblock In {\em Conference of European Association for Machine Translation},
  pp. 261--268, 2012.

\bibitem{ming2017understanding}
Y.~Ming, S.~Cao, R.~Zhang, Z.~Li, Y.~Chen, Y.~Song, and H.~Qu.
\newblock Understanding hidden memories of recurrent neural networks.
\newblock {\em arXiv preprint arXiv:1710.10777}, 2017.

\bibitem{nallapati2016abstractive}
R.~Nallapati, B.~Zhou, C.~Gulcehre, B.~Xiang, et~al.
\newblock Abstractive text summarization using sequence-to-sequence rnns and
  beyond.
\newblock {\em arXiv preprint arXiv:1602.06023}, 2016.

\bibitem{narayanan2018humans}
M.~Narayanan, E.~Chen, J.~He, B.~Kim, S.~Gershman, and F.~Doshi-Velez.
\newblock How do humans understand explanations from machine learning systems?
  an evaluation of the human-interpretability of explanation.
\newblock {\em arXiv preprint arXiv:1802.00682}, 2018.

\bibitem{nguyen2015deep}
A.~Nguyen, J.~Yosinski, and J.~Clune.
\newblock Deep neural networks are easily fooled: High confidence predictions
  for unrecognizable images.
\newblock In {\em Proceedings of the IEEE Conference on Computer Vision and
  Pattern Recognition}, pp. 427--436, 2015.

\bibitem{olah2016attention}
C.~Olah and S.~Carter.
\newblock Attention and augmented recurrent neural networks.
\newblock {\em Distill}, 2016. doi: {{%
10\hspace{.1pt}\discretionary{.}{%
}{.}\hspace{.4pt}23915\discretionary{/}{%
}{/}distill\hspace{.1pt}\discretionary{.}{%
}{.}\hspace{.4pt}00001}}


\bibitem{olah2018building}
C.~Olah, A.~Satyanarayan, I.~Johnson, S.~Carter, L.~Schubert, K.~Ye, and
  A.~Mordvintsev.
\newblock The building blocks of interpretability.
\newblock {\em Distill}, 3(3):e10, 2018.

\bibitem{park2018conceptvector}
D.~Park, S.~Kim, J.~Lee, J.~Choo, N.~Diakopoulos, and N.~Elmqvist.
\newblock Conceptvector: text visual analytics via interactive lexicon building
  using word embedding.
\newblock {\em IEEE transactions on visualization and computer graphics},
  24(1):361--370, 2018.

\bibitem{paulus2017deep}
R.~Paulus, C.~Xiong, and R.~Socher.
\newblock A deep reinforced model for abstractive summarization.
\newblock {\em arXiv preprint arXiv:1705.04304}, 2017.

\bibitem{scikit-learn}
F.~Pedregosa, G.~Varoquaux, A.~Gramfort, V.~Michel, B.~Thirion, O.~Grisel,
  M.~Blondel, P.~Prettenhofer, R.~Weiss, V.~Dubourg, J.~Vanderplas, A.~Passos,
  D.~Cournapeau, M.~Brucher, M.~Perrot, and E.~Duchesnay.
\newblock Scikit-learn: Machine learning in {P}ython.
\newblock {\em Journal of Machine Learning Research}, 12:2825--2830, 2011.

\bibitem{ribeiro2016should}
M.~T. Ribeiro, S.~Singh, and C.~Guestrin.
\newblock Why should i trust you?: Explaining the predictions of any
  classifier.
\newblock In {\em Proceedings of the 22nd ACM SIGKDD International Conference
  on Knowledge Discovery and Data Mining}, pp. 1135--1144. ACM, 2016.

\bibitem{ross2017right}
A.~S. Ross, M.~C. Hughes, and F.~Doshi-Velez.
\newblock Right for the right reasons: Training differentiable models by
  constraining their explanations.
\newblock {\em arXiv preprint arXiv:1703.03717}, 2017.

\bibitem{ruckle2017end}
A.~R{\"u}ckl{\'e} and I.~Gurevych.
\newblock End-to-end non-factoid question answering with an interactive
  visualization of neural attention weights.
\newblock {\em Proceedings of ACL 2017, System Demonstrations}, pp. 19--24,
  2017.

\bibitem{rush2015neural}
A.~M. Rush, S.~Chopra, and J.~Weston.
\newblock A neural attention model for abstractive sentence summarization.
\newblock {\em arXiv preprint arXiv:1509.00685}, 2015.

\bibitem{see2017get}
A.~See, P.~J. Liu, and C.~D. Manning.
\newblock Get to the point: Summarization with pointer-generator networks.
\newblock {\em arXiv preprint arXiv:1704.04368}, 2017.

\bibitem{simonyan2013deep}
K.~Simonyan, A.~Vedaldi, and A.~Zisserman.
\newblock {Deep inside convolutional networks: Visualising image classification
  models and saliency maps}.
\newblock {\em arXiv preprint arXiv:1312.6034}, 2013.

\bibitem{smilkov2017direct}
D.~Smilkov, S.~Carter, D.~Sculley, F.~B. Vi{\'e}gas, and M.~Wattenberg.
\newblock Direct-manipulation visualization of deep networks.
\newblock {\em arXiv preprint arXiv:1708.03788}, 2017.

\bibitem{strobelt2018lstmvis}
H.~Strobelt, S.~Gehrmann, H.~Pfister, and A.~M. Rush.
\newblock Lstmvis: A tool for visual analysis of hidden state dynamics in
  recurrent neural networks.
\newblock {\em IEEE transactions on visualization and computer graphics},
  24(1):667--676, 2018.

\bibitem{sutskever2014sequence}
I.~Sutskever, O.~Vinyals, and Q.~V. Le.
\newblock Sequence to sequence learning with neural networks.
\newblock In {\em Advances in neural information processing systems}, pp.
  3104--3112, 2014.

\bibitem{vaswani2017attention}
A.~Vaswani, N.~Shazeer, N.~Parmar, J.~Uszkoreit, L.~Jones, A.~N. Gomez,
  {\L}.~Kaiser, and I.~Polosukhin.
\newblock Attention is all you need.
\newblock In {\em Advances in Neural Information Processing Systems}, pp.
  6000--6010, 2017.

\bibitem{wu2016google}
Y.~Wu, M.~Schuster, Z.~Chen, Q.~V. Le, M.~Norouzi, W.~Macherey, M.~Krikun,
  Y.~Cao, Q.~Gao, K.~Macherey, et~al.
\newblock Google's neural machine translation system: Bridging the gap between
  human and machine translation.
\newblock {\em arXiv preprint arXiv:1609.08144}, 2016.

\bibitem{xu2015show}
K.~Xu, J.~Ba, R.~Kiros, K.~Cho, A.~Courville, R.~Salakhudinov, R.~Zemel, and
  Y.~Bengio.
\newblock Show, attend and tell: Neural image caption generation with visual
  attention.
\newblock In {\em International Conference on Machine Learning}, pp.
  2048--2057, 2015.

\bibitem{yosinski2015understanding}
J.~Yosinski, J.~Clune, A.~Nguyen, T.~Fuchs, and H.~Lipson.
\newblock Understanding neural networks through deep visualization.
\newblock {\em arXiv preprint arXiv:1506.06579}, 2015.

\bibitem{zeiler2014visualizing}
M.~D. Zeiler and R.~Fergus.
\newblock {Visualizing and Understanding Convolutional Networks}.
\newblock In {\em Computer Vision--ECCV}, vol. 8689, pp. 818--833. Springer,
  2014. doi: {{%
10\hspace{.1pt}\discretionary{.}{%
}{.}\hspace{.4pt}1007\discretionary{/}{%
}{/}978\discretionary{%
}{-}{-}3\discretionary{%
}{-}{-}319\discretionary{%
}{-}{-}10590\discretionary{%
}{-}{-}1\_53}}


\bibitem{Zintgraf17visualizing}
L.~M. Zintgraf, T.~S. Cohen, T.~Adel, and M.~Welling.
\newblock Visualizing deep neural network decisions: Prediction difference
  analysis.
\newblock {\em ICML}, 2017.

\end{thebibliography}
\end{document}